\documentclass[10pt, a4paper]{article}

\usepackage[final]{lrec2026} % this is the new style
\usepackage[utf8]{inputenc}

\usepackage{microtype}

\usepackage{inconsolata}

\usepackage{graphicx}

\usepackage{tabularx}
\usepackage{multirow}
\usepackage{tabularray}

\usepackage[colorinlistoftodos,textsize=tiny,disable]{todonotes}

\usepackage{amsmath}

\usepackage{enumitem}

\usepackage{listings}

\usepackage{tcolorbox}
\usepackage{caption}
\usepackage{booktabs}

\usepackage{graphicx}
\usepackage{amsmath}

\usepackage{lipsum}
\usepackage{booktabs}

\usepackage{tikz}
\usepackage{fontawesome5}

\title{Frame-Guided Synthetic Claim Generation for Automatic Fact-Checking Using High-Volume Tabular Data}

\name{Jacob Devasier, Akshith Putta, Qing Wang, Alankrit Moses, Chengkai Li} 

\address{The University of Texas at Arlington \\
         \{jacob.devasier, cli\}@uta.edu\\}

\abstract{
Automated fact-checking benchmarks have largely ignored the challenge of verifying claims against real-world, high-volume structured data, instead focusing on small, curated tables. We introduce a new large-scale, multilingual dataset to address this critical gap. It contains 78,503 synthetic claims grounded in 434 complex OECD tables, which average over 500K rows each. We propose a novel, frame-guided methodology where algorithms programmatically select significant data points based on six semantic frames to generate realistic claims in English, Chinese, Spanish, and Hindi. Crucially, we demonstrate through knowledge-probing experiments that LLMs have not memorized these facts, forcing systems to perform genuine retrieval and reasoning rather than relying on parameterized knowledge. We provide a baseline SQL-generation system and show that our benchmark is highly challenging. Our analysis identifies evidence retrieval as the primary bottleneck, with models struggling to find the correct data in massive tables. This dataset provides a critical new resource for advancing research on this unsolved, real-world problem.
 \\ \newline  }

\begin{document}

\maketitleabstract

\section{Introduction}\label{sec:intro}
Automatic fact-checking has been widely studied in recent years, both as a means of mitigating misinformation and of identifying hallucinations in large language models. 
Much of this work has focused on using unstructured data, particularly fact-checks from trustworthy sources, both for performing claim matching~\cite{shaar-etal-2020-known, putta-etal-2025-claimcheck} and for training or guiding large language models to generate fact-check verdicts and explanations~\cite{Singhal2024EvidencebackedFC, Khaliq2024RAGARYF, Cheung2023FactLLaMAOI}. 

Structured tabular data from Wikipedia~\cite{chen2020tabfactlargescaledatasettablebased, Aly2021FEVEROUSFE} and scientific documents~\cite{wang-etal-2021-semeval, akhtar-etal-2022-pubhealthtab} has also been utilized to fact-check claims extracted from Wikipedia~\cite{bouziane-etal-2021-fabulous} and real-world claims~\cite{wang-etal-2021-semeval, akhtar-etal-2022-pubhealthtab}. However, these studies exhibit two major limitations: (1) they only consider data that has already been processed and prepared for easy consumption by readers and (2) the volume of individual tables is very small, often fitting entirely into the context window of LLMs. These works' reliance on highly-curated data hinders their ability to fact-check novel claims which do not have clean evidence readily available, and their limited volume doesn't fully evaluate the performance of systems at scale.

\begin{figure}
  \centering
  \includegraphics[width=\linewidth]{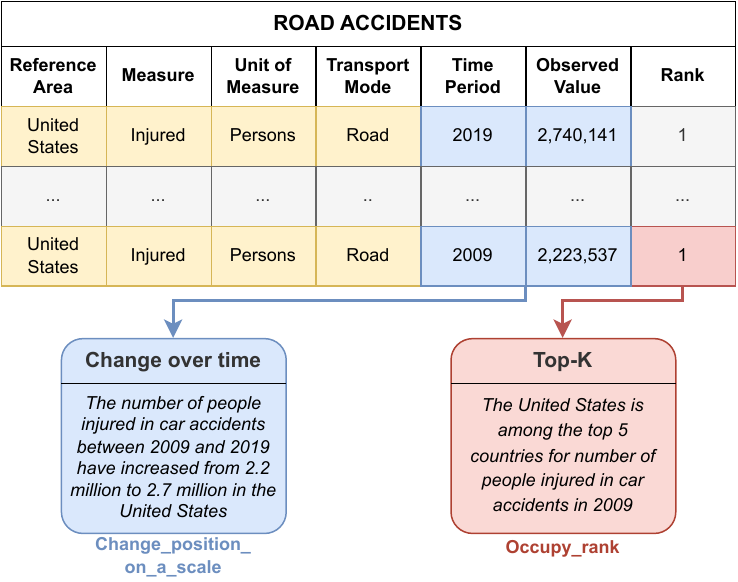}
  
  \caption{\label{fig:claim-types}An example of how our system generates factual claims using specific data from a table. Blue and red colors show the data used to create their corresponding claim. Below each claim is the corresponding semantic frame the claim type is grounded in.}
\end{figure}

One previous study~\cite{claimlens-acl25} explored this direction of using complex, real-world data by collecting factual claims related to OECD (Organization for Economic Co-operation and Development) statistics.~\footnote{\url{https://www.oecd.org/en/data.html}} However, their dataset was limited in size, containing only $\sim$70 claims. This small scale is insufficient for comprehensively evaluating a system's retrieval and reasoning capabilities. In this work, we extend this initial concept to create a much larger-scale dataset of synthetic claims for evaluating automatic fact-checking systems on high-volume, complex structured data. We also expand beyond English to include Chinese, Hindi, and Spanish, enabling the evaluation of multilingual retrieval and reasoning systems. 

Our dataset is generated from a large-scale database of 434 high-volume data tables collected by the OECD. To create claims at scale while ensuring their relevance and complexity, we developed a novel generation methodology. We first identified six common claim types inspired by semantic frames~\cite{arslan-etal-2020-modeling} frequently occurring in fact-checked claims. For each claim type, we designed algorithms to programmatically select significant data points from the tables. This programmatic selection allowed us to generate a large and diverse dataset. These data points were then provided to an LLM, namely Qwen3-225B-A22B, to generate fluent, natural-language factual claims, as shown in Figure~\ref{fig:claim-types}.
\todo[color=green]{Can we use ``corpus'' to descrbie collection tables?}
\todo[color=green]{is the ``filter''-ing explained later?}

This process resulted in a dataset of 78,503 synthetic claims, consisting of roughly 47K English claims and $\sim$10K claims in each of Chinese, Spanish, and Hindi. All claims are mapped to the specific data samples used to create them. To validate the quality of this generation process, we manually reviewed 317 claims and found high quality rates across English (87\%), Chinese (91\%), and Spanish (88\%).

We conducted a novel analysis of the parameterized knowledge within Qwen3 and observed that nearly all specific facts in our dataset are absent from its pre-training corpus. This finding is crucial as a benchmark for this task should measure a system’s ability to retrieve and reason over external evidence rather than rely on memorized knowledge. To verify this, we prompted Qwen3 to predict masked factual values directly from claims using only its internal knowledge. The model achieved an exact match in just 2\% of cases, confirming that successful performance on our dataset requires genuine evidence retrieval and reasoning, not recall of pre-trained facts.

In addition to the dataset, we implemented a baseline system that employs an LLM to generate SQL queries executable on the OECD database. Our baseline leverages the distinct values of each column to construct these queries, substantially reducing the required LLM context length. Experiments show that our new benchmark is highly challenging. State-of-the-art tabular reasoning systems fail almost completely, and our more robust baseline still struggles. An error analysis reveals the primary bottleneck is evidence retrieval—failing to identify the correct table or extract the correct data—which leads to a high rate of "Not Enough Information" predictions.

Our work differs from prior datasets~\cite{chegini-etal-2025-repanda, zhao-etal-2024-findver, Lu2023SCITABACA, Aly2021FEVEROUSFE, chen2020tabfactlargescaledatasettablebased} in several key aspects. First, we manually identify types of claims based on the theory of frame semantics instead of letting an LLM have free reign on tables. Second, we algorithmically select specific data samples for each of these claim types, a process guided by our manual definitions that enables scalable generation. Third, and most importantly, our approach generates claims from large-scale tabular data, with each table averaging around 500K rows---a sharp contrast to prior studies that rely on small, highly curated, Wikipedia-like tables. Finally, we generate multilingual claims from the English-only source data.

% Our work differs from prior datasets~\cite{chegini-etal-2025-repanda, zhao-etal-2024-findver, Lu2023SCITABACA, Aly2021FEVEROUSFE, chen2020tabfactlargescaledatasettablebased} because (1) we manually identify types of claims based on the theory of frame semantics instead of letting an LLM have free reign on tables, (2) we automatically select specific data samples for each type of claim to generate claims,\todo{add information to help understand how such manual work is feasible, although we generated a large number of claims.} (3) most importantly, our approach generates claims from large-scale tabular data, with each table averaging around 500K rows---a sharp contrast to prior studies that rely on small, highly curated, Wikipedia-like tables, and (4) we generate multilingual claims from the English-only data.\todo[color=red]{This paragraph fits Introduction better; how about merge it into the Contribution list at the end of Introduction?}

To summarize, our contributions are as follows:
\begin{itemize}[noitemsep,topsep=2pt,leftmargin=*]
    \item We create and release~\footnote{Data and code can be found at \url{https://anonymous.4open.science/r/oecd-dataset-submission-6780/}} a large-scale, multilingual dataset of 78,503 factual claims mapped to 434 high-volume structured data tables from the OECD.
    \item We propose a novel, frame-guided methodology for generating claims by programmatically selecting significant and meaningful data samples from complex tables.
    \item We conduct a novel analysis of parameterized knowledge, demonstrating that our dataset tests active retrieval and reasoning rather than LLM memorization. \todo[color=green]{It appears this bullet point was not explained with sufficient detail in previous paragraphs.}
    \item We provide a baseline system and a detailed error analysis, identifying evidence retrieval over high-volume tables as a critical and unsolved challenge for future research.
  % \item We propose a novel method for guiding factual claim generation with semantic frames.
  % negative claim generation
  % \item We performed a novel analysis of the parameterized knowledge of OECD data within Qwen3.\todo{It appears this bullet point was not explained with sufficient detail in previous paragraphs. } 
  % \item We create and release a dataset of 78K multilingual factual claims mapped to high-volume structured databases collected from the OECD.
\end{itemize}

\section{Dataset Generation}\label{sec:methodology}
Our dataset generation process is designed to create a large-scale, multilingual fact verification dataset grounded in structured OECD data. The process consists of five main stages. First, in \textit{Data Collection}, we compile a large corpus of statistical tables from the OECD (Section~\ref{subsec:oecd-data-collection}). Second, in \textit{Claim Type Definition}, we define six distinct types of factual claims inspired by common semantic frames evoked in factual assertions (Section~\ref{subsec:claim-generation}). Third, in \textit{Data Selection}, we develop algorithms to programmatically identify and extract significant data samples from the tables that can form the basis for each claim type (Section~\ref{subsec:data-selection}). Fourth, in \textit{True Claim Generation and Curation}, we use a large language model (LLM) to generate natural language claims from these data samples, then employ LLM judges to verify their factual consistency against the source data, producing a clean set of true claims which is subsequently partitioned into train and test splits (Section~\ref{subsec:dataset-curation}). Finally, in \textit{False Claim Generation}, we create a parallel set of false claims by applying systematic, predefined factual perturbations to the true claims (Section~\ref{subsec:false-nei-claims}).

\todo[color=green]{Before we go on to describe the details in each of these subsections, we shall have some form of overview of everything together. Maybe that's what the above paragraph is trying to get at. But the way it is written it is more aligned with the contribution list at the end of Introduction.}

\todo[color=green]{Some of the subsection titles do not match the content well. For instance, ``Claim Generation'' doesn't discuss how claims are generated.}

\subsection{OECD Data Collection}\label{subsec:oecd-data-collection}
We use the same OECD country statistics as \citet{claimlens-acl25} for the source data to create our dataset. These statistics cover a wide range of topics related to health, environment and climate change, finance, employment, agriculture, and many others. This data can be explored using the OECD Data Explorer.~\footnote{\url{https://data-explorer.oecd.org/}} Collecting these statistics resulted in 434 distinct tables with an average of 596,552 rows per table.

\subsection{Claim Type Definition}\label{subsec:claim-generation}
To generate meaningful claims from the OECD data, we first identified six common types of factual claims. The definition of these types is influenced by semantic frames commonly evoked in factual claims, as studied by~\citet{arslan-etal-2020-modeling}, an analysis of common claim structures found on fact-checking platforms. 
\todo[color=green]{The last two sentences are quite confusing. The writing is also rough.  ``largely influenced'', ``commonly evolved'', ``previously studied .... previously studies...'', ``studied for structured data''.} 
Our claim types, which are detailed in Section~\ref{subsec:data-selection}, are based on the frames \textit{Change\_position\_on\_a\_scale}, \textit{Comparing\_at\_two\_different\_points\_in\_time}, \textit{Comparing\_two\_entities}, and \textit{Occupy\_rank}. We excluded the frames \textit{Ratio}, \textit{Recurring\_action}, and \textit{Uniqueness\_of\_trait}, as programmatically identifying interesting and unambiguous data samples for them was particularly challenging. 
% Definitions for all of these frames are provided in Table~\ref{tab:frame-definitions} in the Appendix. \todo{make table}
\todo[color=green]{If we say which are included and which are not, then it appears we have some ``complete set'' to begin with. Where does that complete set come from? Jacob's 2025 paper, Fatma's 2020 paper, or something else? It is not clear and thus not clear why we explain which were excluded. Similarly, hard to understand ``use most of the frames'' --- should be clear these are from fatma's paper now}

A common concern with fact verification datasets is the complexity of the claims. In previous works, this is typically addressed with either multi-hop information retrieval requirements or factual granularity (e.g., part of the claim is false)~\cite{fever-2024-1, Khaliq2024RAGARYF}. In this work, we do not generate highly complex claims or multi-claim statements. We justify this approach by noting that most automatic fact verification systems first decompose complex statements into individual atomic claims before verification~\cite{putta2025claimcheckrealtimefactcheckingsmall, Braun2024DEFAMEDE}.
% We justify this approach as nearly all fact verification systems are capable of breaking down factual claims into individual/atomic claims with very high accuracy~\todo{cite}.

\subsection{Data Selection}\label{subsec:data-selection}
For each claim type, we collected data samples which can be used to generate such claims. 
\todo[color=green]{Is each claim type just corresponding to a frame? If so, we shall make it more explicit, e.g., by calling it ``claim-related frame'' instead of ``claim type''. --- not exactly and not all of them, claim type is generally easier to understand and more accurate}

\subsubsection{Data Preprocessing and Filtering}
% Our objective is to select data from each dataset $d \in D$ that can generate meaningful factual claims. Each dataset $d$ contains a collection of columns $C$, which we categorize into two primary types: metadata columns and measure columns. 
\paragraph{Data structure.} Each OECD table\todo[color=green]{What is a ``dataset''? I thought we only have one dataset?} contains a collection of columns which we categorize into two primary types: metadata columns and measure columns.\todo[color=green]{Is Figure 1 a ``dataset''? We need to use an example to explain the two types of columns. --- yes, and metadata columns are not useful at all, so theres no reason to show them.} Metadata columns contain information shared across nearly all tables, including table names, observation frequencies, and unique identifiers. Measure columns capture the specific measurements conducted within each table. While some measure columns appear frequently across tables (such as measure type, unit of measure, time period, and reference area), others are table-specific (for example, \textit{Transport Mode} in Figure~\ref{fig:claim-types}). Many measure columns also have associated identifier columns; for example, a \textit{Reference Area} column with values like "Sweden" might have an associated \textit{Reference Area ID} column with codes like "SWE".
% Many measure columns include associated identifier columns that provide codes corresponding to plain text values, such as the reference area identifier \textit{SWE} for \textit{Sweden}.
\todo[color=green]{This sentence is obscure. Is it something like key-foreign key reference? We can use that concept to explain it. Is it about two columns in the same table or across 2 different tables? Again, a more comprehensive example to illustrate all these concepts will be useful.}\todo[color=green]{This paragraph is not about data selection itself; rather, it is about types of columns. It might make more sense to discuss this elsewhere, such as an overview of the dataset. --- it has necessary prerequisite info for next paragraphs}

\paragraph{Preprocessing.} Our data selection process begins with table preprocessing. We first identify and remove metadata columns and measure identifier columns from consideration. We eliminate any columns or rows that contain entirely null values. We remove rows where the primary observation value (the numerical data point found in the \textit{obs\_value} column) is null or where the \textit{observation status} is not classified as \textit{normal} (e.g., 'Estimated' values for future dates). Tables lacking \textit{Reference Area} (e.g., country) information are also excluded.
\todo[color=green]{What are ``observed values''? Are they a subset of ``measure columns''? How to determine which columns are observed values?}\todo[color=green]{Ideally you would need some examples for explaining some of the listed observation statuses. --- not very important as they are metadata stuff, added some examples that could be there}

\paragraph{Time-window filtering.} To ensure claims are generated from time periods with robust and comparable data, we identify an "ideal time window" for each table. We first find the time period with the maximum number of reporting countries (the "maximum coverage"). We then select time periods where the country count is within 5\% of this maximum. This "leniency" allows for small, temporary dips in reporting countries (e.g., due to different reporting frequencies like quarterly vs. annual) within an otherwise stable time window. Tables that do not contain at least a two-year ideal time window and include at least 20 countries are excluded.
% To ensure we only generate claims from time periods where there are a significant number of other countries,\todo{``other'' than what?} we identify an ``ideal time window'' for each dataset by finding periods where the number of countries represented in each time period falls within 5\% of the maximum coverage,\todo{What ``maximum coverage''?} allowing up to four time periods of leniency.
\todo[color=green]{cannot understand ``time periods of leniency'' and the whole sentence.}

\paragraph{Measure combinations.} Following preprocessing, our cleaned tables typically contain reference area, time period, and a collection of measure columns, along with the primary observation value. To analyze the data, we retrieve the unique combinations of non-numeric measure column values. For example, a table might contain three measure columns: \textit{Measure}, \textit{Unit of Measure}, and \textit{Transport Mode}. We identify all occurring unique combinations of these values, such as (\texttt{Injury crash}, \texttt{Crashes}, \texttt{Road}) or (\texttt{Fatalities}, \texttt{Persons}, \texttt{Road}).

Each unique combination corresponds to a specific perspective for generating claims about the table (e.g., \textit{the number of crashes where someone was injured in a road accident} for (\texttt{Injury crash}, \texttt{Crashes}, \texttt{Road})). We refer to these column-value combinations as \textit{measure combinations}.

% Following the preprocessing phase, our cleaned tables typically contain reference area information, time period data, and a collection of measure columns. We retrieve unique combinations of column values that represent different ways to interpret and analyze the data. For example, the table shown in Figure~\ref{fig:claim-types} contains three measure columns: \textit{Measure}, \textit{Unit of Measure}, and \textit{Transport Mode}. We identify 3 unique combinations of these values: \texttt{Injury crash-Crashes-Road}, \texttt{Fatalities-Persons-Road}, and \texttt{Injured-Persons-Road}.\todo{What do you mean by ``unique combinations''? How about other combinations? I gues they don't have satisfying rows?} 

\subsubsection{Claim-Specific Data Extraction} Next, we implement six distinct data extraction strategies, one for each claim type. These strategies are applied within each measure combination to find significant data points.
% Next, we implement six distinct claim type-specific data extraction strategies, each designed to capture different types of factual relationships within the data.\todo{The last sentence is disconnected from the previous sentences which were about measure combinations.} 

\todo[color=teal]{To explain these claims, you need some running example data table; and use real claims in addition to the pattern that mentions X Y Z. --- i think it is fine now without running examples, but maybe in the future we can improve this}

\paragraph{Top-K claims.}
These claims identify countries ranking at the extremes for a measure, following the pattern: ``Country X is among the top/bottom 5 countries on measure M in year Y.'' \textbf{Logic:} For each measure combination and year, we rank countries based on their \textit{observed value}. We select the top and bottom $k$ samples. We set $k=5$ for subsets with >50 countries (for that measure combination and year) and $k=3$ for subsets with 20--50 countries. Subsets with <20 countries are excluded.
% Top-k claims identify countries that rank among the highest or lowest performers for specific measures in given years, following the pattern ``Country X is among the top/bottom 5 countries on measure M in year Y.'' For each measure combination, we select the top and bottom $k$ samples, where $k$ is determined by the subset size: we use $k=5$ for subsets with more than 50 samples, $k=3$ for subsets with 20--50 samples, and exclude subsets with fewer than 20 samples from top-k claim generation.
\todo[color=teal]{If M is a measure combination, you cannot rank the countries with a linear order, unless you use a ranking function, or something like skyline layers. --- lets discuss this later} 
\todo[color=green]{The notion of ``subset'' is not clear.}

\paragraph{Constant-Change claims.}
These claims capture sustained trends, following the pattern: ``Country X has shown a constant increase/decrease on measure M for N consecutive years, as of year Y.'' \textbf{Logic:} We identify continuous spans of at least eight years showing a consistent increase or decrease in the \textit{observed value} for a country within a measure combination. In the claim, $N$ represents this identified span (e.g., $N=8$, $9$, ...). We simply select the start and end data points of this span for each claim.
\todo[color=green]{The current pattern is not natural. It will have to be something like ``has increased for N consecutive years till year Y.'' --- the pattern isn't necessarily supposed to be english. it is just a template sentance kinda. the LLM will build a fluent sentence integrating the those key values.}
\todo[color=teal]{again, if it is a measure combination, you cannot say increase or decrease. In fact, all the claim types appear to have this problem.}
\todo[color=green]{If it is always ``eight years'', we won't want to say ``N years'' at the same time.}
\todo{Not sure why you mention ``start and end data points''. The pattern doesn't mention such points.}

\paragraph{Historical-Extreme claims.}
These claims identify locally unprecedented performance, following the pattern: ``In year Y, Country X recorded its highest/lowest value on measure M in the last N years.'' \textbf{Logic:} For each country and measure combination, we analyze its time-series data to find the maximum number of years ($N$) since a higher (or lower) value was recorded. We require $N \ge 10$ years to ensure the claim represents a notable event.
% Historical-extreme claims identify when countries achieve locally unprecedented performance levels, following the pattern ``Country X is highest/lowest on measure M in N years in year Y.'' For each country and measure combination, we analyze time-series data to determine the maximum number of years since a higher or lower value was recorded. We require at least ten years to have elapsed since the previous extreme value, to ensure the claim represents a notable achievement or decline.
\todo[color=green]{Again, the pattern doesn't sound like natural English. --- see above}

\paragraph{Change-in-Rank claims.}
These claims document significant shifts in relative performance, following the pattern: ``Country X went from rank A to rank B on measure M between year Y and year Z.'' \textbf{Logic:} We identify instances where countries experience substantial changes in their relative rankings over time. A change is ``substantial'' if it is: (1) a flat change of at least 10 positions or 20\% of the total number of countries in that subset, whichever is larger; or (2) a change-in-rank ratio of at least 2 for the same country (e.g., moving from rank 10 to rank 5).
% We examine time-series data for each reference area across all measure combinations and identify instances where countries experience substantial changes in their relative rankings compared to other countries over time. We consider substantial changes to be either: (1) flat changes of either 10 positions or 20\% of the total number of countries or (2) a change ratio of at least 2 between the highest and lowest ranks.
\todo[color=green]{Earlier, you mentioend ``subset'' to refer to the applicable countries. It seems you will need to say something similar here, instead of ``total number of countries''.}

\paragraph{Change-Over-Time claims.}
These claims focus on absolute value changes, following the pattern: ``Country X went from value A to value B on measure M between year Y and year Z.'' \textbf{Logic:} These claims use the same underlying data points as the Change-in-Rank claims but emphasize the change in observed values rather than positional changes in rankings.
% Change-over-time claims focus on absolute value changes rather than relative rankings, following the pattern ``Country X went from value A to value B on measure M between year Y and year Z.'' These claims utilize the same underlying data as change-in-rank claims but emphasize the change in observed values rather than positional changes in rankings.

\paragraph{Have-Trait claims.}
These claims provide simple factual statements, following the pattern: ``Country X has value A on measure M in year Y.'' \textbf{Logic:} We generate these claims exclusively using data from the individual temporal data points (e.g., the start and end points) identified for the Change-in-Rank and Change-Over-Time claim types, creating separate atomic claims for each time point.
% Have-trait claims provide simple factual statements about country characteristics at specific points in time, following the pattern ``Country X has value A on measure M in year Y.'' We also generate these claims using data from change-in-rank; however, we create separate claims for each temporal data point to capture country-specific characteristics across different time periods.
\todo[color=green]{Other than data from change-in-rank, what other data were used in generating have-trait claims? --- none, maybe the "also" caused this confusion} 
\todo{Does the order of claim types in the explanations follow any order?}

\subsection{Claim Generation and Curation}\label{subsec:dataset-curation}
Our data selection process (Section~\ref{subsec:data-selection}) yielded 104,930 data samples that met our extraction criteria (e.g., being a top-k value, a historical extreme, etc.). To avoid biasing our dataset towards tables with especially high volume or claim types which are more abundant, we limited the number of samples drawn from each OECD table for each claim type, aiming to generate a maximum of 100 claims in English and 20 in each non-English language.
\todo[color=green]{what ``significance threshold''? --- extraction criteria}

We generated claims for each data sample using Qwen3-225B-A22B. We faced two
primary concerns with using LLM-generated outputs to build a dataset. First, the factuality of claims with respect to the intricate and detailed facts of the world are difficult to ensure. Second, it may be difficult to communicate a fact clearly without sometimes losing subtle details. This can lead to or be exacerbated by low linguistic quality of the sentence. 

To address the factual consistency of the claims, we used two LLM judges (Llama 3.3 70B and Qwen3-225B-A22B) to verify that the generated claim was factually supported by the provided data sample (detailed in Section~\ref{subsec:llm-judge}). To address linguistic quality, we relied on Qwen3's thinking mechanism, which allows the model to refine its output rather than generating it in a single pass. In total, our dataset consists of 87,517 claims determined to be factually consistent by both LLM judges.

\begin{table}[t]
\centering
\resizebox{\linewidth}{!}{
\begin{tabular}{lrrrr}
\toprule
\textbf{Claim Type} & \textbf{English} & \textbf{Chinese} & \textbf{Hindi} & \textbf{Spanish} \\
\midrule
Change-in-Rank      & 10,350 & 2,208 & 2,207 & 2,123 \\
Have-Trait          & 9,089 & 1,897 & 1,913 & 1,882 \\
Change-Over-Time    & 8,928 & 1,988 & 2,004 & 1,974 \\
Top-K               & 8,827 & 2,012 & 1,969 & 1,860 \\
Constant-Change     & 5,093 & 1,784 & 1,606 & 1,410 \\
Historical-Extreme  & 4,544 & 908 & 933 & 925 \\
\midrule
Total & 46,831 & 10,797 & 10,632 & 10,174 \\
\bottomrule
\end{tabular}
}
\caption{Number of samples in our dataset broken down by language and claim type.}\label{tab:language_task_counts}
\end{table}

\paragraph{Data partitioning.} 
To build useful train and test partitions, we ensured our test set would evaluate out-of-domain performance by holding out at least 10\% of the OECD tables entirely from the training set. This partitioning process resulted in a final training set of 75,666 claim-data pairs and a test set of 2,837 claim-data pairs (totaling 78,503 claims). The remaining 9,014 claims (from the original 87,517) were set aside to ensure a clean separation between in-domain and out-of-domain tables. We sampled roughly 10\% of these test set claims for human evaluation (Section~\ref{subsec:human-annotation}). The language and claim type counts for the final dataset are shown in Table~\ref{tab:language_task_counts}.

% To build useful train and test partitions, we wanted to ensure our test set has the following features: (1) it should contain at least 1 sample of every dataset, language, and claim type combination; (2) at least 10\% of the datasets are excluded from the training set to evaluate out-of-domain performance; (3) there is no overlap in data samples between the train and test sets.\todo{It seems by definition the data samples are distinct. Is there anything we need to do to ``ensure'' there is no overlap?} This led to 9,014 training samples being removed from the training set, resulting in a final training set of 75,666 data-claim pairs and a test set of 2,837 data-claim pairs, totaling 78,503 samples. We sample roughly 10\% of these test samples for human evaluation in Section~\ref{subsec:human-annotation}.
\todo[color=green]{It appears earlier you were calling each chosen data point (roughly speaking) a ``data sample''. Now you are calling a ``data-claim pair'' a sample.}

\subsection{Creating False Claims}\label{subsec:false-nei-claims}
A fact verification dataset requires both true and false claims. Having generated a set of verified true claims (Section~\ref{subsec:dataset-curation}), we created a corresponding set of false claims by applying predefined factual perturbations to the original true claims. We define four types of perturbations based on the claim type: \textbf{Numeric} perturbations, where the given value is multiplied by a randomly selected scaling factor (e.g., 0.5, 1.5, 2.0); \textbf{Rank} perturbations, where the rank is randomly shifted by a substantial amount (e.g., moving a rank to a much higher/lower position); \textbf{Duration} perturbations, where the duration in years is randomly altered, for instance by shifting a start/end year by 2-6 years or extending a duration by 3-8 years; and \textbf{Binary} perturbations, where the claim's direction is inverted (e.g., \textit{increased} → \textit{decreased}, or \textit{top-K} → \textit{bottom-K}).

\todo[color=green]{The notion of ``producing false claims'' comes as a surprise. This was not discussed and as a reader I was not prepared to see this. Need to explain why we want to produce false claims.}
\todo{Also, if we need false claims anyways, why did we discuss earlier ways to avoid false claims? Need to clarify.}

% To produce claims which the given data are unable to fact-check, we define another set of perturbations which can be applied. These perturbations are as follows: out-of-range value perturbation, where we change either the country or year to be one not in the range of values for the measure; claim specificity perturbation, where we introduce additional qualifiers to claims such that there is no data available to fact-check it. 

% Two additional perturbations were considered for NEI samples, but we felt were signficant challenges to ensure were not actually checkable. These are out-of-domain measures, where we would change a measure to one not in the data, and claim ambiguity, where we remove necessary information to make a claim uncheckable. Ensuring either of these types of perturbations do not result in checkable claims would require significant manual effort, so we leave these as potential future directions.

\section{Baseline System}\label{sec:baseline}
We implement a baseline system to extract evidence and perform reasoning over large, structured tables. We store the OECD Data tables in a SQL database, for ease of storage and retrieval. Prior approaches~\cite{chen2020tabfactlargescaledatasettablebased, Lu2023SCITABACA} typically serialize entire tables into text sequences provided to a language model within its input context. However, this design quickly becomes infeasible for large tables, as the serialized representation can exceed the context length limits of even modern LLMs. Our baseline instead decomposes the problem into modular retrieval and reasoning steps that avoid such scaling bottlenecks.

Given a claim $c$, the system first uses an LLM to decompose it into a set of atomic subclaims $c_1, \ldots, c_k$, where the number of subclaims is determined dynamically by the LLM at inference time using in-context learning. 
% Listing \ref{lst:subclaim_prompt} details how we prompt the LLM to generate multiple decomposed subclaims.\todo[color=green]{how is the decomposition done. We shall say a few words and include references if we use some known methods.}\todo[color=green]{How is the number determined?} 
This atomic fact decomposition enables the system to retrieve heterogeneous evidence---potentially from different tables or from multiple regions within a single table---that would not be captured by a single SQL query.\todo[color=green]{It is surprising to see SQL. It has never been mentioned and its role is unclear. -- Have mentioned it above in 1st paragraph.}

For each subclaim, the system retrieves the most semantically similar table using the \texttt{gte-multilingual-base} embedding model~\footnote{\url{https://huggingface.co/Alibaba-NLP/gte-multilingual-base}} and its reranker.\footnote{\url{https://huggingface.co/Alibaba-NLP/gte-multilingual-reranker-base}}\todo{say a few words about how the embedding model and reranker are used in retrieving semantically similar tables. Include references if you are using known methods.} To support this search, we pre-compute textual representations for all tables in the OECD dataset. Each table representation concatenates the table name, the OECD-provided description, and the names and representative values of categorical data columns. These representations are embedded using the aforementioned models and form the searchable corpus for table retrieval. The table with the highest cosine similarity to a subclaim's embedding is selected for further processing.
\todo[color=green]{We haven't said the ``representations'' are embeddings.}

Once a relevant table is identified, the system prompts an LLM to generate an executable SQL query that extracts the evidence needed to verify the subclaim. The prompt includes the claim text, table name, OECD description, column names, and a subset of unique categorical values for each column.%, as detailed in Listing~\ref{lst:sql_gen_prompt}.
\footnote{The \textit{obs\_value} column is excluded from the prompt.}\todo[color=green]{Have we mentioned the \textit{obs\_value -- We mention it in 2.3.1, preprocessing section} column somewhere? Otherwise its mentioning in the footnote is sudden.} For columns with more than 20 unique values, we use BM25 retrieval to select the 20 most similar values relative to the subclaim. The SQL generation step is retried up to three times if execution fails, similar to the SQL generation in~\citet{claimlens-acl25}.\todo[color=green]{Are we using some approach from the prior work? provide references.}\todo{provide an example with the corresponding SQL query.}

After retrieving the relevant data, the system prompts the LLM again %(Listing~\ref{lst:subclaim_verification}) 
to determine whether each subclaim is \textit{True}, \textit{False}, or \textit{Not Enough Information (NEI)} based on the query results. Finally, these subclaim-level judgments are synthesized into an overall verdict for the original claim using another LLM call.% (see Listing~\ref{lst:subclaim_synthesis}).

\section{Experiments}\label{sec:experiments}

\todo[color=red]{Which experiments are for verifying the usefulness of the dataset, as motivated in Introduction? I couldn't locate such experiments.}
\subsection{Human-Annotated Quality Check}\label{subsec:human-annotation}
To evaluate the quality of the LLM-generated synthetic claims, we manually annotated 317 claims randomly selected from the test set, including 143 English, 66 Chinese, 59 Spanish, and 49 Hindi claims.
Each language was annotated by a native speaker of the  respective language. Each annotator was instructed to label the data-claim pair as ``good'' or ``bad'' and provide a comment for any bad claims. %The exact instructions given to the participants is shown in Listing~\ref{lst:annotation_instructions} in the appendix. 

We found similar quality rates across English (87\%), Chinese (91\%), and Spanish (88\%) with each being rated ``good'' around 90\% of the time. Hindi, however, had a much lower rate of only 66\%. Based on the annotator's comments, we found that 13\% of the Hindi samples had only minor issues related to the use of ``among OECD countries'' within the claim which we later determined to be acceptable.
\todo[color=green]{Is this about Hindi or all languages? Unclear what the ``issue'' is. Particularly, ``within the claim'' is confusing.}

\begin{table}
\centering
\resizebox{\linewidth}{!}{
\begin{tabular}{c c c c c }
\multicolumn{2}{c}{} & \multicolumn{3}{c}{\textbf{Llama 3.3}} \\% \cline{2-5}
&           & \textbf{True}  &  \textbf{False} & \textbf{NEI}      \\
\multirow{3}{*}{\rotatebox{90}{\textbf{Qwen3}}}
& \textbf{True}      & 87,517 (84.2\%) & 579 (0.6\%) & 4,695 (4.5\%) \\
& \textbf{False}     & 4,679 (4.5\%) & 441 (0.4\%) & 339 (0.3\%) \\
& \textbf{NEI}       & 1,683 (1.6\%) & 33 (0.0\%) & 3914 (3.8\%) \\
\end{tabular}
}
\caption{Distribution of samples predicted to be True, False, or NEI by two LLM-based factuality evaluators.}\label{tab:llm-evaluation}
\end{table}
\subsection{LLM-as-a-Judge Factuality Check}\label{subsec:llm-judge}
We also performed a more robust evaluation (Table~\ref{tab:llm-evaluation}) of the factuality of the entire dataset using two different LLM judges: Qwen3-225B-A22B and Llama 3.3 70B. We chose to include Llama 3.3 as a judge to mitigate bias that occurs between models within the same family~\cite{Li2025PreferenceLAA,Panickssery2024LLMERA}. Both LLMs were prompted 
% (Listing~\ref{lst:llm_judge}) 
to verify the factual correctness of the generated claims based on the data sample and the dataset description. Each claim was classified as True, False, or Not Enough Information (NEI), and was accompanied by a justification generated by the LLM. Llama 3.3 frequently had trouble with exact value mismatches (e.g., it would classify a claim as False if any numeric rounding was involved), so we adjusted the prompt to allow for minor rounding differences. The distribution of the judged outputs from this experiment is shown in Table~\ref{tab:llm-evaluation}. 

We found that 16\% of the generated claims were not predicted to be True by at least one of the models.\todo[color=green]{16\% seems incorrect. There are NEIs. --- 84\% true, rest not true} Constant-Change claims exhibited the lowest proportion of claims predicted True by both (75\%).\todo[color=green]{Is 75\% meant to be the ``true positive rate''? Its description doesn't match the definition of true positive rate.} Upon examining a sample of 100 claims, we observed that this was typically caused by the way the data was presented to the LLM. A similar trend was observed for other claim types with Llama 3.3, where many predictions cited insufficient supporting data, leading to a relatively high overall rate of Not Enough Information (NEI) predictions (8.6\%).\todo[color=yellow]{Are you saying the low true positive rate and the NEI are the same issue? It is confusing. What is the ``trend''? what is causing the high NEI rate? --- the data was passed to llama, but it didn't realize that it was passed all it needed, so it said NEI} We also found that most cases in which Qwen3 predicted a claim as False were due to minor rounding differences in numerical values. Such variations were permitted during claim generation to make the claims sound more natural.

\begin{table}[t]
    \centering
    \resizebox{0.9\linewidth}{!}{
    \begin{tabular}{llc}
        \toprule
        \textbf{Model} & \textbf{Tables} & \textbf{Accuracy (\%)} \\ 
        \midrule
        TabSQLify           & Predicted & 3.9 $\pm$ 1.3 \\ % 0.031746, 0.047619, 0.025397, 0.041270, 0.050794
        Baseline no-think   & Predicted & 11.3 $\pm$ 2.0 \\
        Baseline            & Predicted & 15.5 $\pm$ 0.7 \\
        \midrule
        TabSQLify           & Predicted-HQ & 3.8 $\pm$ 1.5 \\
        Baseline no-think   & Predicted-HQ & 14.9 $\pm$ 2.6 \\
        Baseline            & Predicted-HQ & \textbf{20.4 $\pm$ 2.6} \\
        \midrule
        TabSQLify           & Gold      & 6.4 $\pm$ 1.5 \\ % 0.066667, 0.076190, 0.044444, 0.069841, 0.063492
        Baseline no-think   & Gold      & 26.4 $\pm$ 2.3 \\
        Baseline            & Gold      & 36.0 $\pm$ 0.8 \\
        \midrule
        TabSQLify           & Gold-HQ   & 6.4 $\pm$ 1.6 \\
        Baseline no-think   & Gold-HQ   & 32.9 $\pm$ 2.0 \\
        Baseline            & Gold-HQ   & \textbf{42.7 $\pm$ 5.9} \\
        \bottomrule
        % \multicolumn{3}{c}{* Uses Qwen3-30B-A3B and 15k context length (original is 3k) for fairness}
        % \multicolumn{3}{*}{* Uses Qwen3-30B-A3B and 15k context length (original is 3k) for fairness}
    \end{tabular}
    }
    \caption{Accuracy of our baseline system on the human evaluation dataset. HQ indicates that we excluded samples marked as bad by humans.}
    \label{tab:end-to-end}
\end{table}

\subsection{Baseline Performance}\label{subsec:baseline-performance}\todo[color=red]{When we call it ``Baseline'', the reader will expect there is another more sophisticated solution. But it appears we only experimented with ''baseline''?}
In this experiment, we compare our baseline system (Section~\ref{sec:baseline}) with TabSQLify~\cite{nahid2024tabsqlifyenhancingreasoningcapabilities}, a state-of-the-art system in tabular reasoning, on the human-evaluated subset of the test set. TabSQLify originally used GPT-3.5 with a context length of 3K tokens, so to ensure a fair comparison we reimplemented their system with Qwen3-30B-A3B with a 15K token context length. This comparison is presented in Table~\ref{tab:end-to-end}. 

As we expected, TabSQLify failed to handle the high-volume tables in the OECD dataset, resulting in an accuracy of 6.4\% on the gold tables and 3.9\% when using the tables predicted by our baseline.\todo[color=green]{how to interpret ``predicted by our baseline'' and ''gold'' tables? How is Baseline's accuracy on Predicted (by baseline) only 15\%? --- very difficult} Our baseline performed significantly better on both predicted and gold tables, with the thinking-enabled LLM consistently performing the best, with 36\% and 20.4\% accuracy on the gold and predicted tables, respectively. 
After removing low-quality (``bad'') test samples identified by human annotators (Section~\ref{subsec:human-annotation}), we observed a significant improvement in our baseline's performance, coupled with a larger variance between runs. However, this improvement was not observed for TabSQLify, suggesting that its failures are unrelated to the quality of the claims.

% \input{tables/baseline_evaluation} 
% Probably just gonna comment this out. I don't think we really need to discuss this. We can add it to the appendix and discuss more in the camera ready version.

% \input{tables/subclaim-accuracy}
% Same with this table and analysis.

\begin{table*}[t]
    \centering
    % \resizebox{\linewidth}{!}{
    \begin{tabular}{lccccc}
        \toprule
        \textbf{Predicted Verdict} & \textbf{Overall} & \textbf{English} & \textbf{Spanish} & \textbf{Chinese} & \textbf{Hindi}\\% & \textbf{Verdict Accuracy} & \textbf{\# Samples}\\
        \midrule
        \multicolumn{5}{l}{\textit{Baseline table retrieval accuracy}} \vspace{.5mm} \\
        Overall & 32.2\% & 36.6\% & 33.5\% & 29.3\% & 30.0\% \\%& 29.7\% & 2,837 \\
        True    & 75.0\% & 78.6\% & 77.9\% & 77.8\% & 67.2\% \\%& 90.9\% & 265 \\
        False   & 36.8\% & 40.1\% & 38.0\% & 33.8\% & 35.6\% \\%& 56.5\% & 1,066 \\
        NEI     & 21.3\% & 22.7\% & 24.8\% & 19.5\% & 18.6\% \\%& 0.0\% & 1,505 \\
        \midrule
        \multicolumn{5}{l}{\textit{Baseline data retrieval accuracy}} \vspace{.5mm} \\
        Overall  & 18.3\% & 22.8\% & 18.5\% & 14.9\% & 17.3\% \\%& 29.7\% & 7,902 \\
        True     & 52.2\% & 52.7\% & 54.7\% & 55.2\% & 48.0\% \\%& 90.9\% & 706  \\
        False    & 21.0\% & 24.9\% & 21.7\% & 18.6\% & 19.1\% \\%& 56.5\% & 3,123  \\
        NEI      & 10.4\% & 13.2\% & 11.8\% &  6.9\% & 10.0\% \\%& 0.0\% & 4,073 \\
        \bottomrule
    \end{tabular}
    % }
    \caption{Subclaim-level evidence retrieval performance on table retrieval (top block) and data retrieval (bottom block). The Verdict Accuracy column shows the accuracy of the predicted verdicts, duplicated for both retrieval blocks. The overall accuracy refers to the weighted average of all the verdicts.}
    \label{tab:evidence-accuracy}
\end{table*}

We also evaluated our system's ability to retrieve the correct evidence from the database, as shown in Table~\ref{tab:evidence-accuracy}. For this experiment, we used the entire test set to obtain more robust metrics.
\todo[color=green]{several issues in Table caption: 1) it is not the ``right-most column''; 2) What is ``Predicted Accuracy'' (also, in the table it says ``verdict accuracy''); 3) what is ``supplid Category counts''. 4) What does ``Overall'' accuracy mean? It is macro-average (across langauge and across True/False/NEI)? --- seems like this was addressed?}
Overall, we found that the \texttt{gte-multilingual} semantic similarity models performed well across all language for both evidence retrieval tasks (table retrieval and data retrieval), though the models performed noticeably worse on Hindi table retrieval than the other languages for the True verdict predictions.
\todo[color=green]{The comment regarding Hindi is only applicable for table retrieval?}
\todo[color=teal]{need correct/incorrect signals in the table to finish this analysis @akshith}

As expected, the table retrieval accuracy is highest for verdicts predicted as True. This is also the least predicted verdict, as the LLM seems to only predict True when it is more confident. The primary source of error for our baseline appears to be NEI predictions, which account for 53\% of all predictions. False claims appear to pose less of a challenge, as the LLM can predict claims to be false using contradicting evidence that does not necessarily originate from the same data as the claim.\todo[color=green]{``false verdicts'' means predicted as False, or the verdict is wrong? If the former, maybe write it as ``False verdicts''.}\todo[color=green]{at least couple of grammar errors.}

\subsection{Parameterized Knowledge Leakage}\label{subsec:param-knowledge}
In this experiment, we aim to analyze the internal knowledge of Qwen3-30B-A3B to understand how much of our baseline's performance can be attributed to the LLM's internal knowledge obtained during its pretraining. Previous studies~\cite{almeida2025tiebe, mousavi-etal-2024-dyknow, WorldBench2024} have also evaluated the knowledge recall abilities of LLMs in this domain; however, the facts they used were very common and therefore more likely to be known by the LLMs. We design two experiments using the Have-Trait claims from our dataset's test set. We choose this claim type due to its simplicity and higher likelihood of the LLM retrieving the corresponding fact. 

\begin{table}[t]
\centering
%\resizebox{\linewidth}{!}{
\begin{tabularx}{\linewidth}{Xcc}
\toprule \textbf{Tolerance Level} & \textbf{Accuracy (\%)} & \textbf{$\Delta$} \\
\midrule
$p=0$       & 2.0   & {--}    \\
$p=0.25$    & 14.9  & $645$\% \\
$p=0.5$     & 24.4  & $64$\%  \\
$p=1.0$     & 36.1  & $48$\%  \\
$p=1.5$     & 42.3  & $17$\%  \\
$p=2.0$     & 47.1  & $11$\%  \\
% $p=0$       & 2.0   & -     \\
% $p=0.25$    & 14.9  & +12.9 \\
% $p=0.5$     & 24.4  & +9.5  \\
% $p=1.0$     & 36.1  & +11.7 \\
% $p=1.5$     & 42.3  & +6.2  \\
% $p=2.0$     & 47.1  & +4.8  \\
\bottomrule
\end{tabularx}
%}
\caption{Accuracy of masked fact predictions across different tolerance levels.}\label{tab:masked-fact-prediction}
\end{table}

\paragraph{Masked fact prediction.}
First, we test whether the specific facts in our dataset are directly represented within the model parameters. Intuitively, if an LLM encodes a fact, it should be able to predict the corresponding value in that fact. To evaluate this, we mask the numerical value of the measure within each claim and prompt the model to predict the missing value. 

We measure the deviation between the LLM’s predicted value and the ground truth value and define a relative tolerance parameter $p$ that sets the acceptable range of values to be between [$\frac{v}{1 + p}$, $v(1 + p)$],\todo[color=green]{$v(1 - p)$ perhaps makes more sense. Just checking whether this was a typo. If indeed $\frac{v}{1 + p}$ was used, it is too late to change right now.} where $v$ is the ground truth value. As shown in Table~\ref{tab:masked-fact-prediction}, the LLM is able to predict the masked value exactly ($p=0$) for roughly 2\% of the claims. We also observe a rapid increase in performance between $p=0$ and $p=0.5$, indicating that the LLM often approximates the correct answers; however, it is evident that most specific facts are not precisely represented within the model's internal knowledge.

\begin{table}
    \small
    \centering
    \begin{tabular}{c c c c}
        \multicolumn{1}{c}{} & & \multicolumn{2}{c}{\textbf{Predicted}} \\
        & & \textbf{True} & \textbf{False} \\
        \multirow{2}{*}{\rotatebox{90}{\textbf{Label}}}
        & \textbf{True}  & 996 (6.8\%) & 6,838 (46.7\%) \\
        & \textbf{False} & 614 (4.2\%)  & 6,195 (42.3\%) \\
    \end{tabular}
    \caption{Confusion matrix for knowledge-claim alignment.}
    \label{tab:alignment-with-generalized-knowledge}
\end{table}

\paragraph{Claim-knowledge alignment.}
Second, we test whether the model’s world knowledge is consistent with the generated claims. To do this, we prompt the LLM with a given claim and directly ask it to predict whether the claim is \textit{True} or \textit{False}. As shown in the confusion matrix in Table~\ref{tab:alignment-with-generalized-knowledge}, the LLM has a heavy bias toward predicting a claim to be \textit{False} (89\%) regardless of its truthfulness. The LLM achieves a precision of 61.8\%, a recall of 12.7\%, and an F1-score of 21.1\%.

% \begin{table}[t]
% \centering
% \resizebox{\linewidth}{!}{
% \begin{tabular}{lrrrr}
% \toprule
% \textbf{Tolerance Level} & \textbf{TT} & \textbf{TF} & \textbf{FT} & \textbf{FF} \\
% \midrule
% p=0      & 21(0.39\%) & 87(1.61\%) & 674(12.5\%) & 4612(85.5\%) \\
% p=0.25   & 247(4.58\%) & 556(10.31\%) & 448(8.31\%) & 4143(76.81\%) \\
% p=0.5    & 337(6.99\%) & 937(17.37\%)& 318(5.9\%)& 3762(69.74\%)\\
% \bottomrule
% \end{tabular}
% }
% \caption{.}
%     \label{tab:combine-task1-task2}
% \end{table}
\begin{table}[t]
\centering
\resizebox{0.9 \linewidth}{!}{
\begin{tabular}{cc|cc}
\toprule
\multicolumn{2}{c}{\textbf{Correct Predictions}} & \multicolumn{2}{c}{\textbf{Tolerance Level}} \\
% \cmidrule(lr){2-3}
Task 1 & Task 2 & \textbf{$p = 0$} & \textbf{$p = 0.1$}  \\
\midrule  
\faCheck & \faCheck   & 0.4\%  & 2.5\%  \\
\faCheck & \faTimes   & 1.6\%  & 5.0\%   \\
\faTimes & \faCheck   & 12.5\% & 10.4\%   \\
\faTimes & \faTimes   & 85.5\% & 82.1\%    \\

% Task 1 (\faCheck) $\cap$ Task 2 (\faCheck)   & 21 (0.4\%) & 132 (2.5\%)   \\
% Task 1 (\faCheck) $\cap$ Task 2 (\faTimes)   & 87 (1.6\%)& 269 (5.0\%)  \\
% Task 1 (\faTimes) $\cap$ Task 2 (\faCheck)   & 674 (12.5\%) & 563 (10.4\%)      \\
% Task 1 (\faTimes) $\cap$ Task 2 (\faTimes)   & 4,612 (85.5\%) & 4,430 (82.1\%) \\
\bottomrule
\end{tabular}
}
\caption{Consistency of LLM predictions between masked fact prediction (Task 1) and alignment with generalized knowledge (Task 2) across different tolerance levels.}
    \label{tab:combine-task1-task2}
\end{table}

To determine whether the LLM genuinely possesses the knowledge required to correctly predict the masked values in the claims---rather than arriving at them by chance---we analyze the consistency of its predictions across these two experiments.
To do this, we only consider the true claims, as predicting a false claim to be \textit{False} does not necessarily entail the fact is known by the LLM. The results are presented in Table~\ref{tab:combine-task1-task2}. 

We found that only 0.4\% of the claims in our test set were correctly predicted ($p=0$) in both of these tasks, meaning that the LLM has likely only been exposed to a few of the claims in our dataset during its pretraining. At $p=0.1$, this increases to 2.5\%. Based on these results, we conclude that nearly all of the performance from our baseline can be attributed to the system itself rather than any internal knowledge held by the LLM.

\section{Related Works}\label{sec:background}\todo[color=green]{If won't make much sentence to say ''Background'', as this section is at the end.}

\paragraph{Table-to-text generation.}
\citet{zhao-etal-2023-investigating} investigated the table-to-text generation capabilities of large language models (LLMs) in real-world information-seeking scenarios. Their analysis concluded that open-source LLMs still lag behind GPT-4 in generating text from tables; however, the gap between open and closed-source models has closed significantly since their work was published. 
Many previous works~\cite{zhang-etal-2024-opent2t, Iravani2024TowardsME, Min2024ExploringTI, Sundararajan2024ImprovingFA, Gong2019TabletoTextGW} have studied table-to-text generation. However, these approaches typically grant LLMs excessive freedom in text generation, often producing outputs that are overly general, verbose, or unfocused with respect to a specific fact. In contrast, our work aims to generate concise, fact-specific statements grounded in data. To achieve this, we guide the LLM to produce claims that align with semantic frames commonly observed in real-world fact-checked claims~\cite{arslan-etal-2020-modeling}.

\paragraph{Fact verification with tabular data.} 
FEVEROUS~\cite{Aly2021FEVEROUSFE} is a fact verification dataset using Wikipedia tables and text as evidence. The claims in FEVEROUS were generated by taking sentences from Wikipedia articles and mutating them to create false claims. TabFact~\cite{chen2020tabfactlargescaledatasettablebased} (and RePanda~\cite{chegini-etal-2025-repanda}, an extension of TabFact) is another table-based fact verification dataset using Wikipedia tables. FINDVER~\cite{zhao-etal-2024-findver} is a recent benchmark consisting of long financial documents with many tables in the documents. SCITAB~\cite{Lu2023SCITABACA} is another table-based fact verification dataset focusing on scientific tables. However, all of these datasets use relatively small tables (less than 100 rows on average) and do not consider high-volume data as we do in this work.

\paragraph{Querying high-volume data with LLMs.}
\citet{claimlens-acl25} is the primary motivation for this work. They collected a dataset of real-world factual claims from trustworthy fact-checking sources and used OECD data to fact-check them. However, their dataset was very limited in size (roughly 70 claims) due to the difficulty of collecting real-world claims from the OECD datasets. In this work, we instead generate synthetic claims which can be fact-checked using OECD data, allowing us to create a much larger dataset. \citet{radhakrishnan2024knowingaskbridging} also built a system to integrate high-volume data with LLMs to improve factual accuracy of responses to user's statistical queries. They chose to use Data Commons, a large collection of public datasets from trusted organizations, much like the OECD datasets. This work trained an LLM to convert natural language queries into Data Commons queries which were then mapped to a set of templated queries to retrieve relevant data. One limitation of their work is that they do not query their data with SQL, which is a widely studied task with LLMs.

% https://arxiv.org/pdf/2305.14987
% Investigating Table-to-Text Generation Capabilities of LLMs in Real-World Information Seeking Scenarios
% Tabular data is prevalent across various industries, necessitating significant time and effort
% for users to understand and manipulate for their
% information-seeking purposes. The advancements in large language models (LLMs) have
% shown enormous potential to improve user efficiency. However, the adoption of LLMs in
% real-world applications for table information
% seeking remains underexplored. In this paper, we investigate the table-to-text capabilities
% of different LLMs using four datasets within
% two real-world information seeking scenarios.
% These include the LOGICNLG and our newlyconstructed LOTNLG datasets for data insight
% generation, along with the FeTaQA and our
% newly-constructed F2WTQ datasets for querybased generation. We structure our investigation around three research questions, evaluating
% the performance of LLMs in table-to-text generation, automated evaluation, and feedback
% generation, respectively. Experimental results
% indicate that the current high-performing LLM,
% specifically GPT-4, can effectively serve as a
% table-to-text generator, evaluator, and feedback
% generator, facilitating users’ information seeking purposes in real-world scenarios. However, a significant performance gap still exists between other open-sourced LLMs (e.g.,
% TÜLU and LLaMA-2) and GPT-4 models.

\section{Conclusion}\label{sec:conclusion}
In this work, we address a significant gap in automated fact-checking by moving beyond small, curated tables to high-volume, complex structured data. We introduced a novel, frame-guided methodology to generate a large-scale, multilingual dataset of 78,503 synthetic claims. Grounded in massive OECD data tables, our dataset is not only substantial in size but also diverse, covering English, Chinese, Spanish, and Hindi, and is structured around six common semantic frames to ensure claims are realistic and varied.

Our experiments validate the challenge and utility of this new benchmark. We demonstrated through knowledge-probing experiments that the facts in our dataset are not typically memorized by LLMs, meaning systems must perform genuine evidence retrieval and reasoning rather than relying on parameterized knowledge. Our baseline system, which decomposes claims and generates SQL queries, struggled to perform well on our system (9.3\% on the full test set). Our error analysis revealed the primary bottleneck is evidence retrieval---specifically, failing to find the correct table or extract the correct data, leading to a high rate of ``Not Enough Information'' predictions.

The difficulty our baseline faced demonstrates that fact-checking on large-scale, complex structured data is a non-trivial and largely unsolved and underexplored problem. This dataset provides a valuable new resource for the community to develop and benchmark more robust models, highlighting a critical need for future research into improved table retrieval and precise data extraction techniques.
This dataset provides a valuable new resource for the research community to develop and benchmark more robust models, underscoring the critical need for future work on improved table retrieval and precise data extraction techniques.

\todo[color=green]{Are limitations and ethics still required? --- not for lrec}

\bibliographystyle{lrec2026-natbib}
\bibliography{lrec2026-example}

% \appendix

% \input{sections/appendix}

\end{document}